\documentclass[letterpaper, 10 pt, conference]{ieeeconf}

\IEEEoverridecommandlockouts
\usepackage{cite}

\usepackage[pagewise]{lineno}
\usepackage[ruled,vlined]{algorithm2e}
\usepackage{amsmath}
\usepackage{amssymb}
\SetKwComment{Comment}{/* }{ */}

\DeclareMathOperator*{\argmin}{arg\,min}

\usepackage{graphicx}
\usepackage{subcaption}
\usepackage{xcolor}

\graphicspath{{./figures/}}

\title{\LARGE \bf
Adaptive Planning and Control with Time-Varying Tire Models for Autonomous Racing Using Extreme Learning Machine
}
\author{Dvij Kalaria$^{1}$, Qin Lin$^{2}$, and John M. Dolan$^{1}$
\thanks{$^{1}$The authors are with the Robotics Institute, Carnegie Mellon University {\tt\small dkalaria@andrew.cmu.edu},{\tt\small jdolan@andrew.cmu.edu}}
\thanks{$^{2}$Qin Lin is with the Electrical Engineering and Computer Science Department, Cleveland State University {\tt\small q.lin80@csuohio.edu}}
}

\begin{document}

\maketitle
\thispagestyle{empty}
\pagestyle{empty}

\begin{abstract}

Autonomous racing is a challenging problem, as the vehicle needs to operate at the friction or handling limits in order to achieve minimum lap times. Autonomous race cars require highly accurate perception, state estimation, planning and precise application of controls. What makes it even more challenging is the accurate identification of vehicle model parameters that dictate the effects of the lateral tire slip, which may change over time, for example, due to wear and tear of the tires. Current works either propose model identification offline or need good parameters to start with (within 15-20\% of actual value), which is not enough to account for major changes in tire model that occur during actual races when driving at the control limits. We propose a unified framework which learns the tire model online from the collected data, as well as adjusts the model based on environmental changes even if the model parameters change by a higher margin. We demonstrate our approach in numeric and high-fidelity simulators for a 1:43 scale race car and a full-size car.

\end{abstract}


\section{INTRODUCTION}

The estimation of tire conditions is critical for the performance and safety of racing. Racing teams usually pre-estimate tire models and tire traction to assist in decision-making on when to change tires \cite{WEST202014456}, and what tire to use for minimal cumulative lap times to win the race. Race drivers adjust their behaviors with each lap, adjusting speeds at the turns based on their estimates of how well the tire traction will support them. Estimating the vehicle's handling capacity is particularly challenging due to a lot of variables around the actual race such as temperature, weather conditions, etc.  For example, the tire forces depend on the racing surface and temperature. 

To build an autonomous racing car, a model-based planning and control framework is commonly used, assuming precise vehicle model parameters are available. While estimating the kinematics of the vehicle is pretty straightforward, as it only requires calculating geometric parameters of the vehicle, estimating precise dynamics, especially lateral tire forces, requires interacting with the road material and hence requires time-intensive system identification and verification. We propose an online learning strategy that first uses a geometric-based kinematic nominal model \cite{JainBayesRace2020} and identifies the complicated dynamics such as tire model, aerodynamic forces, and longitudinal friction from the collected data even under environmental changes due to wear and tear of the tires. This significantly reduces the effort required in model identification and allows online adaptation of racing line reference speeds to adjust the driving behavior accordingly. Specifically, we propose the use of an Extreme Learning Machine (ELM)-based tire force prediction model for lateral forces and longitudinal tire model, which captures the discrepancy between the nominal model and the dynamic model. Our approach using ELMs separately for rear and front tires allows interpreting the coefficient friction of the tires, which can be used for adaptive speed planning along the racing line in contrast to existing approaches assuming time-invariant friction. Our approach significantly improves the safety of a race car. First, we demonstrate our results in a 1:43-scale numeric simulator \cite{Liniger2015OptimizationbasedAR}, where we change the tire coefficient of friction to simulate wear and tear of the tires during the run. We then demonstrate in the high-fidelity Carla simulator with additional sensor and model noise, and a sloped racetrack. 

The main contributions of our work are summarized as follows: 

\begin{itemize}
\item We propose a novel unified optimization framework for model identification, as well as online adaptation for autonomous racing. Our approach outperforms the state-of-the-art Gaussian process (GP)-based controller \cite{JainBayesRace2020} on lap time, safety, and computation efficiency.
\item Unlike existing approaches that assume time-invariant friction, our ELM model estimates time-varying friction online to enable adaptive speed planning along the race line for safe racing.
\item We demonstrate results in a 1:43 numeric simulator and Carla with simulated wear and tear of the tires.
\end{itemize}




\section{RELATED WORKS}
\label{sec:related}

The task of autonomous racing has been actively explored both in the optimal control and reinforcement learning communities. 
The existing approaches can be briefly classified into model-free and model-based categories. Some model-free works, e.g.,  \cite{Wadekar2021TowardsED, Pan2017AgileOA, Cai2021VisionBasedAC} train end-to-end policies to directly take images as input and output throttle and steering. 

In the model-based category, \cite{Liniger2015OptimizationbasedAR} and \cite{Kabzan2019AMZDT} propose modular frameworks for autonomous racing using uncertainty-aware nonlinear Model Predictive Control (MPC) with accurate tire model parameters.

MPC can be implemented either in a receding horizon manner or a contouring manner (called MPCC \cite{Liniger2015OptimizationbasedAR}). The high performance of MPC relies on accurate dynamic models. As a remedy, learning-based control algorithms are expected to robustly reject disturbances caused by inaccurate dynamic models. In light of this, \cite{Rosolia2019LearningHT} \cite{JainBayesRace2020} propose an iterative procedure that uses data from previous laps to correct the vehicle model. \cite{Hewing2017CautiousNW} proposes using GP and linear MPC \cite{8920657} to enable learning vehicle parameters. Model mismatch errors of up to 15\% can be compensated. All these methods assume the availability of nearly-accurate tire parameters, which is practically difficult.

\cite{JainBayesRace2020} proposes using an extended kinematic model and applies GP to learn the difference between the high-fidelity dynamic model and the extended kinematic model from collected data offline. However, using GP in optimization is computationally expensive (0.25s-0.3s according to \cite{JainBayesRace2020}), which makes it practically infeasible to deploy on an actual car. Our work is closest to \cite{JainBayesRace2020}, but it is importantly able both to learn the vehicle model and adapt online. Both of these advantages are realized by using ELMs, which are computationally lightweight (0.02s-0.06s per iteration), to estimate tire model instead of GP.

The rest of the paper is organized as follows:  Section \ref{sec:preliminaries} briefly defines some preliminaries used in the work. Section \ref{sec:method} elaborates on our proposed method. Section \ref{sec:results} reports the simulation experiments. Finally, in Section \ref{sec:conclusion} we make concluding remarks and discuss some interesting future directions. 


\section{PRELIMINARIES} \label{sec:preliminaries}

\subsection{Racing Line} \label{subsec:racing_line}

During a race, professional drivers follow a racing line while using specific maneuvers that allow them to use the limits of the car’s tire forces. This path can be used as a reference by the autonomous racing motion planner to effectively follow time-optimal trajectories while avoiding collision, among other objectives. The racing line is essentially a minimum-time path or a minimum-curvature path. They are similar but the minimum-curvature path additionally allows the highest cornering speeds given the maximum legitimate lateral acceleration \cite{doi:10.1080/00423114.2019.1631455}. 

In our work, we calculate the minimum-curvature optimal line, which is close to the optimal racing line as proposed by \cite{doi:10.1080/00423114.2019.1631455}. The race track information is input by a sequence of tuples ($x_i$,$y_i$,$w_i$), $i \in \{0,...,N-1\}$, where ($x_i$,$y_i$) denotes the coordinate of the center location and $w_i$ denotes the lane width at the $i^{th}$ point. The output trajectory consists of a tuple of seven variables: coordinates $x$ and $y$, longitudinal displacement $s$, longitudinal velocity $v_x$, acceleration $a_x$, heading angle $\psi$, and curvature $\kappa$. The trajectory is obtained by minimizing the following cost:
\begin{equation} \label{opt_racing_line_eqn}
\begin{split}
    & \mathop{\min}\limits_{\alpha_1...\alpha_N} \quad \sum_{n=0}^{N-1} \kappa_i^2(n)\\
    \text{s.t.} & \quad \alpha_i \in \left[ -w_i+\frac{w_{veh}}{2},w_i-\frac{w_{veh}}{2} \right]\\
\end{split}
\end{equation}
where the vehicle width is $w_{veh}$, and $\alpha_i$ is the lateral displacement at the $i^{th}$ position with respect to the reference center line. To generate a velocity profile, the vehicle's velocity-dependent longitudinal and lateral acceleration limits are required  \cite{doi:10.1080/00423114.2019.1631455}. We generate velocity profiles for different lateral acceleration limits determined based on the coefficients of friction of the front tire ($\mu_f$) and rear tire ($\mu_r$), with mass $m$ and the gravity constant $g$, as described next. 

\begin{figure}[htbp]
\begin{subfigure}{.295\textwidth}
    \centering
    \includegraphics[width=\textwidth]{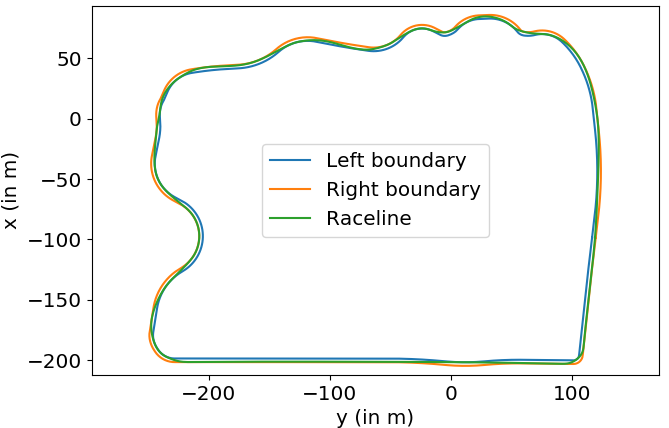}
    \caption{}
    \label{fig:racing_line_carla}
\end{subfigure}
\begin{subfigure}{.185\textwidth}
    \centering
    \includegraphics[width=\textwidth]{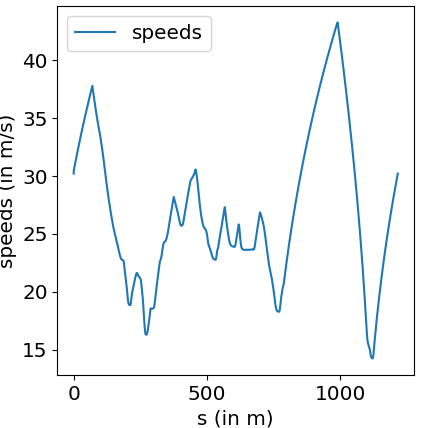}
    \caption{}
    \label{fig:racing_line_speeds_carla}
\end{subfigure}
\caption{Racing line (Carla track) and speed profile. The $x$ axis of the speed profile is the longitudinal displacement.}
\label{fig:racing_line_carla_}
\end{figure}

\begin{figure}[htbp]
\begin{subfigure}{.28\textwidth}
    \centering
    \includegraphics[width=\textwidth]{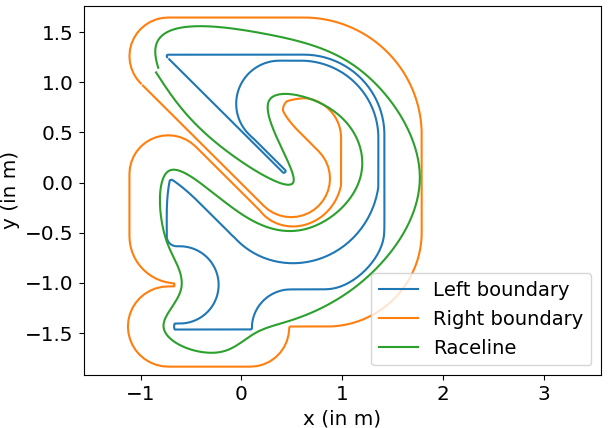}
    \caption{}
    \label{fig:racing_line_ethz}
\end{subfigure}
\begin{subfigure}{.20\textwidth}
    \centering
    \includegraphics[width=\textwidth]{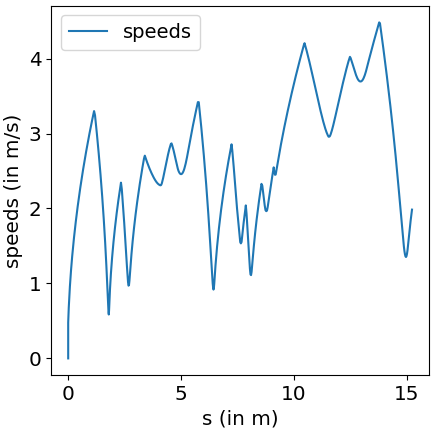}
    \caption{}
    \label{fig:racing_line_speeds_ethz}
\end{subfigure}
\caption{Racing line (ETHZ track) and speed profile. The $x$ axis of the speed profile is the longitudinal displacement.}
\label{fig:racing_line_ethz_}
\end{figure}

We define multiple velocity profiles for a range of maximum lateral forces from $\mu_{eff}$ in $[\mu_{min}, \mu_{max}]$ so that we can obtain the velocity at any given values of $\mu_l$ and $\mu_r$ by interpolating between the known values. The racing line and velocity profiles at $\mu_{eff}=1$ are shown for the two tracks used in our experiments, see Fig. \ref{fig:racing_line_carla_} and Fig. \ref{fig:racing_line_ethz_}.

\vspace{-4mm}
\vspace{-6mm}

\vspace{-3mm}
\subsection{Vehicle models}

The most widely used vehicle models in the context of autonomous driving and racing are the kinematic and dynamic bicycle models. The kinematic model uses $x$, $y$ coordinates, $\phi$ heading angle in the global inertial frame, velocity $v$, and acceleration $a$. The dynamic model has states $x$, $y$, and $\phi$ in the global frame; longitudinal velocity $v_x$, lateral velocity $v_y$, and yaw angular velocity $\omega$ in the vehicle's body frame; and steering angle $\delta$. For the kinematic model, the control variables are acceleration $a$ and steering $\delta$, and throttle $d$ and change in steering $\Delta \delta$ define the action space. For the dynamic model, $F_{r,x}$ is the longitudinal force on the rear tire in the tire frame assuming a rear-driven vehicle, $F_{f,y}$ and $F_{r,y}$ are the forces on the front and rear tires, respectively, and $\alpha_f$ and $\alpha_r$ are the corresponding slip angles. We denote the mass of the vehicle $m$, the moment of inertia in the vertical direction about the center of mass of the vehicle $I_z$, the length of the vehicle from the COM (center of mass) to the front wheel $l_f$, and the length from the COM to the rear wheel $l_r$. $B_{f/r}$, $C_{f/r}$, $D_{f/r}$ are the Pacejka tire model parameters specific to the tire and track surface. Just to avoid confusion, these are used only in numeric simulation to simulate the vehicle dynamics; our goal is to estimate the tire curve (slip angle vs. lateral force generated by the tire) using ELMs, as we discuss later. For longitudinal force, $C_{m1}, C_{m2}$ are known constants obtained from the gear model and $C_r, C_d$ are aerodynamic force constants which are learned from vehicle interactions.

\subsubsection{Kinematic model}

This model is defined exclusively by the geometric parameters of the vehicle that can be physically measured, without taking into consideration any contact surfaces. It ignores the effect of lateral tire slip, which is crucial at high speeds. Hence, it is suitable for lower speeds, but at cornering speeds, it is inaccurate due to considerable lateral tire slip. 
\vspace{-2mm}
\begin{equation} \label{eqn:kin_model}
    \begin{bmatrix}
    \dot{x} \\
    \dot{y} \\
    \dot{\phi} \\
    \dot{v} \\
    \dot{\delta} \\
    \beta
\end{bmatrix}
= \begin{bmatrix}
    v \cos (\phi+\beta)\\
    v \sin (\phi+\beta)\\
    \frac{v \sin (\beta)}{l_r} \\
    a \\
    \frac{\Delta \delta}{\Delta t} \\
    \tan ^{-1} (\frac{l_f}{l_f+l_r} tan \delta)
\end{bmatrix} 
\end{equation} 
\vspace{-2mm}
\subsubsection{Dynamic model}

The dynamic model, on the other hand, is complex and takes into consideration the lateral tire slip but requires time-intensive identifications to determine the Pacejka tire model parameters of the tire, drivetrain, and aerodynamic resistance parameters. It is highly suitable for autonomous racing, as it accurately models the dynamics at cornering speeds during turns. However, accurate values of parameters are needed, and these values may change over time. 
Also, the tire lateral slip curves must be re-calibrated if the surface changes, which is common for autonomous racing. Here, $p$ and $r$ are the roll and pitch respectively of the vehicle which are assumed to be numerically less in magnitude such that $\cos(p) \approx 1$ and $\cos(r) \approx 1$

\vspace{-3mm}
\begin{equation} \label{eqn:dyn_model}
\begin{split}
&\begin{bmatrix}
    \dot{x} \\
    \dot{y} \\
    \dot{\phi} \\
    \dot{v}_x \\
    \dot{v}_y \\
    \dot{\omega} \\
    \dot{\delta} \\
\end{bmatrix}
= \begin{bmatrix}
    v_x \cos (\phi) - v_y \sin (\phi)\\
    v_x \sin (\phi) + v_y \cos (\phi) \\
    \omega \\
    \frac{1}{m} (F_{r,x} - F_{f,y} \sin(\delta) + m v_y \omega - m g \sin(p))\\
    \frac{1}{m} (F_{r,y} + F_{f,y} \cos(\delta) - m v_x \omega + m g \sin(r))\\
    \frac{1}{I_z} (F_{f,y} l_f \cos(\delta) - F_{r,y} l_r) \\
    \frac{\Delta \delta}{\Delta t} \\
\end{bmatrix} \\ 
\end{split}
\end{equation}
where $F_{r,x} = (C_{m1} - C_{m2} v_x) d - C_r - C_d v_x^2$, $F_{f,y} = D_f \sin(C_f \tan^{-1}(B_f \alpha_f)), \alpha_f = \delta - \tan^{-1}\left(\frac{\omega l_f + v_y}{v_x}\right)$, and $F_{r,y} = D_r \sin(C_r \tan^{-1}(B_r \alpha_r)), \alpha_r = \tan^{-1}\left(\frac{\omega l_r - v_y}{v_x}\right)$.


\subsubsection{Extended kinematic model}

The essential difference between the two models is that the dynamic model has additional states $v_x$, $v_y$ and $w$ which are not in the kinematic model. Also, slip angles $\alpha_f$ and $\alpha_r$ are assumed to be $0$ in the kinematic model. We therefore consider a variant of the kinematic model called the extended kinematic or E-kinematic model \cite{JainBayesRace2020}. It is derived just by having $\dot{\alpha}_f = 0$ and $\dot{\alpha}_r = 0$. Also, for longitudinal forces, the effect of aerodynamic forces is ignored in this model. Hence, we have:

\begin{equation} \label{eqn:ekin_model}
\begin{split}
& \begin{bmatrix}
    \dot{x} \\
    \dot{y} \\
    \dot{\phi} \\
    \dot{v}_x \\
    \dot{v}_y \\
    \dot{\omega} \\
    \dot{\delta} \\
\end{bmatrix}
= \begin{bmatrix}
    v_x \cos (\phi) - v_y \sin(\phi)\\
    v_x \sin (\phi) + v_y \cos(\phi)\\
    \omega \\
    \frac{F_{r,x}}{m}\\
    \frac{l_f}{l_f + l_r} (\dot{\delta} v_x + \dot{v_x} \delta)\\
    \frac{1}{l_f + l_r} (\dot{\delta} v_x + \dot{v_x} \delta)\\
    \frac{\Delta \delta}{\Delta t} \\
\end{bmatrix} 
\end{split}
\end{equation}
where  $F_{r,x} = (C_{m1} - C_{m2} v_x) d$.

This model also has only geometric parameters $l_f$, $l_r$, and $m$ as model parameters. Unlike the dynamic model, it doesn't consider tire forces. It extends the dimension of the kinematic model to obtain the same number of states as the dynamic model. In our work, the E-kinematic model will serve as a \textbf{nominal model}. The difference between the dynamic model and the nominal model will be learned and compensated by our learning component - ELM.


\subsection{Extreme Learning Machine}

\begin{figure}[htbp]
    \centering
    \includegraphics[width=0.4\linewidth]{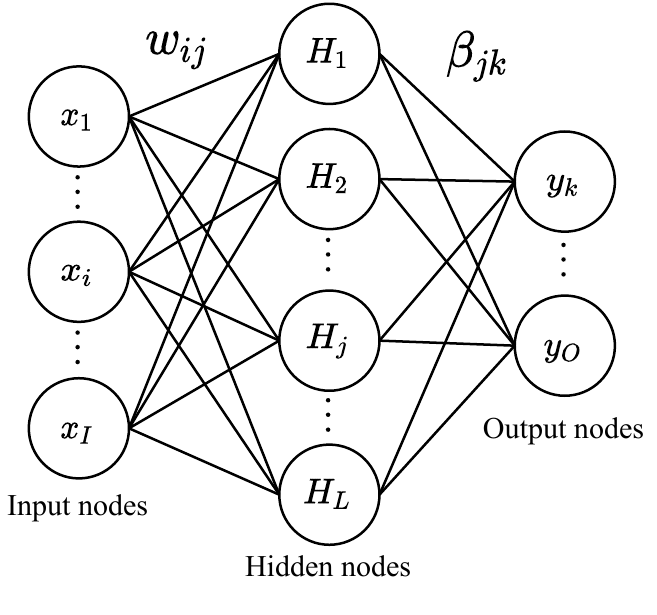}
    \caption{Architecture of a single layer ELM.}
    \label{fig:elm}
\end{figure}

For modeling tire friction curves, we use extreme learning machines (ELMs). An architecture of a single-layer ELM is shown in Fig. \ref{fig:elm}. Let the no of input nodes be $n_i$, no of hidden nodes $n_h$ and no of output nodes $n_o$. Let the input be $x \in R^{n_i}$, then the output $y \in R^{n_o}$ can be obtained as:

\begin{equation}
\mathbf{y}=\mathbf{H}( \mathbf{x}) \boldsymbol{\beta} =  g(\mathbf{w}^T \mathbf{x})\boldsymbol{\beta}
\end{equation}
where $\mathbf{H} = [H_1, H_2, \cdots, H_j, \cdots, H_{n_L}]$ is the non-linear transformation, $\mathbf{w} = [w_1, \cdots, w_{n_I}]$ are input weights, and $\boldsymbol{\beta} = [\beta_1, \cdots, \beta_{n_L}]$ are output weights. Furthermore, $g(\cdot)$ corresponds to a kernel function that fulfills the condition for universal approximation \cite{huang2006universal}.

For training, let us consider the dataset to be of $N$ size with each element of the form $(\mathbf{x}_n, \mathbf{t}_n)$, where $\mathbf{t}_i \in \mathbb{R}^{n_O}$ is the true label corresponding to  $i^{th}$ datapoint. ELM has a fixed input weight denoted as $\hat{\mathbf{w}}$ unlike a traditional feedforward single layer neural network. Using ELM instead of a single layer neural network is more efficient as according to \cite{huang2006extreme}, if we initialize input weight $\mathbf{w}$ with random values drawn from a distribution and keep them fixed, it can theoretically get the same results even if it is not optimized. The learning objective of ELM is to obtain the optimal $\boldsymbol{\beta}^*$ which satisfies the following minimization objective:
\begin{equation}
\label{eq:elm-opt}
 \boldsymbol{\beta}^*= \underset{\boldsymbol{\beta}}{\argmin} \; \mathcal{L}\left(\mathbf{y}(\boldsymbol{\beta}), \mathbf{t}\right)= \underset{\boldsymbol{\beta}}{\argmin} \; \| \mathbf{H}( \mathbf{x}) \boldsymbol{\beta} - \mathbf{t}) \|_2
\end{equation}
where $\mathcal{L}$ is the loss function defined as $l^2$-norm between the output $y$ and the label $t$. We can solve for $\boldsymbol{\beta}^*$ analytically using the following relation 

\begin{equation}
\label{eq:pseudo_inverse}
 \boldsymbol{\beta} = \mathbf{H}^+ \mathbf{t}
\end{equation}

Where $\mathbf{H}^+$ is the pseudo inverse of $\mathbf{H}$. However, such standard analytical solution is possible only when the output labels $\mathbf{t}$ are directly available or if the dataset is fixed. In our case, we use separate ELMs for rear and front tire slip-curve estimation. Here the inputs are the slip angle $\alpha_{f|r}$ and unity $1$ to add bias on the first linear layer, thus $n_I = 2$ and the output is the lateral force, $F_{[f|r]y}$, thus $n_o=1$. $n_h$ is a hyperparameter which we choose as $40$ for the experiments. We will discuss in Sec. \ref{subsec:online_adaptation} why we use a gradient variant of the standard ELM called tuning ELM \cite{ertuugrul2014detailed}, where the gradient signal to reduce the model error is passed through both front and rear tire ELMs. 

\section{METHODOLOGY} \label{sec:method}

\subsection{Problem setting}

We first perform experiments on the 1:43-scale numeric autonomous racing platform \cite{Liniger2015OptimizationbasedAR}, where the dynamic model with the Pacejka tire model is used. We  then also demonstrate our results on the more complicated and close-to-real Carla simulator, where the simulator vehicle model is unknown, but we have estimates of the engine like throttle and rpm-to-torque mapping that can be used to get approximate values of $C_{m1}$ and $C_{m2}$ to use in the e-kinematic model (nominal model $f_{e-kin}$). Later, through the collected data, we learn the residual error $f_{err}$ to obtain the final corrected model, i.e.,  $f_{corr} = f_{e-kin} + f_{err}$. The learnable parameters here are the ELM tire slip models, longitudinal friction $C_r$ and aerodynamic force constant $C_d$.

\begin{figure}[htbp]
    \centering
    \includegraphics[width=0.8\linewidth]{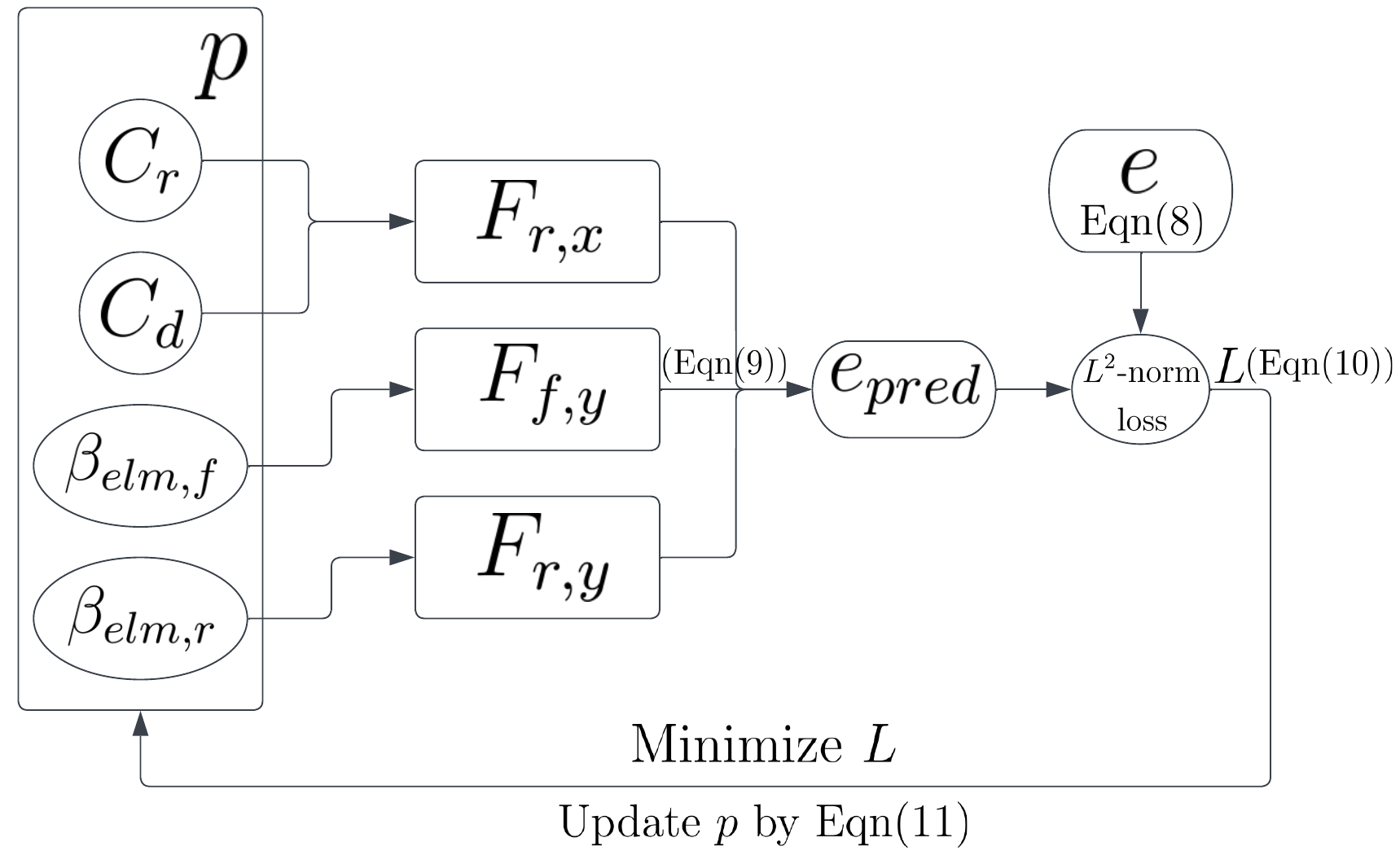}
    \caption{Training cycle}
    \label{fig:train_cycle}
\end{figure}

Our goal is to learn $f_{corr}$ from the vehicle interactions to make real-time decisions to minimize lap time and also adapt to the changing environment, which is simulated in the experiments by reducing the coefficient of friction of the tires and adjusting the speeds accordingly for safety. 

\subsection{Learning based control} 

We break down our approach into three parts as follows:

\subsubsection{Collecting data} 
The collected data can be represented as a pair of states from consecutive time steps, i.e., as $\{x_k,u_k,x_{k+1}\}$. Hence, the error between the actual state $x_{k+1}$ and the one estimated from the discretized e-kinematic model can be represented as:

\begin{equation}
e_{(x_k,u_k)} = x_{k+1} - f_{e-kin} (x_k,u_k)     
\end{equation}

The e-kinematic vehicle model is discretized using Runge-Kutta $6^{th}$-order approximation. We collect the data online, hence the dataset at the $n^{th}$ time step $D_n=\{(e(x_k,u_k))|k \in \{1,2,..,n-1\}\}$. 

\subsubsection{Training and online adaptation} \label{subsec:online_adaptation}
Our goal is to learn the model mismatch error, $e(x_k,u_k)$. The training cycle can be briefly visualized as in Fig. \ref{fig:train_cycle}. The predicted value $e_{pred}(x_k,u_k)$ is obtained as follows:
\vspace{-1mm}
\begin{equation}
\begin{split}
&\begin{bmatrix}
    e_{\dot{v}_x} \\
    e_{\dot{v}_y} \\
    e_{\dot{\omega}} \\
\end{bmatrix}_{pred}
= \begin{bmatrix}
    \frac{1}{m} (- C_r - C_d v_x^2 - F_{f,y} \sin(\delta) + m v_y \omega)\\
    \frac{1}{m} (F_{r,y} + F_{f,y} \cos(\delta) - m v_x \omega)\\
    \frac{1}{I_z} (F_{f,y} l_f \cos(\delta) - F_{r,y} l_r)
     \\
\end{bmatrix} \\
& \ \ \ \ \ \ \ \ - \begin{bmatrix}
     m \sin(p)\\
    - m \sin(r) + \frac{l_f}{l_f + l_r} (\dot{\delta} v_x + \dot{v_x} \delta)\\
    \frac{1}{l_f + l_r} (\frac{\Delta \delta}{\Delta t} v_x + \dot{v_x} \delta)\\
\end{bmatrix} \\ 
\end{split}
\end{equation}
where $F_{f,y} = ELM_f(\alpha_f)$, $F_{r,y} = ELM_r(\alpha_r)$. 

The loss to be minimized is defined as:
\begin{equation} \label{eqn:loss}
\begin{split}    
    &L = \sum_{k \in \{i,..,j\}} \bigg((e_{\dot{v}_x}(x_k,u_k)-e_{\dot{v}_x,pred}(x_k,u_k))^2 \\
    &\ \ \ \ \ \ \ \ \ \ \ \ + (e_{\dot{v}_y}(x_k,u_k)-e_{\dot{v}_y,pred}(x_k,u_k))^2\\ 
    &\ \ \ \ \ \ \ \ \ \ \ \ + (e_{\dot{\omega}}(x_k,u_k)-e_{\dot{\omega},pred}(x_k,u_k))^2\bigg)
\end{split}
\end{equation}

We minimize the loss $L$ iteratively using stochastic gradient descent with momentum. Letting the update rate be $\alpha$ and the momentum be $\gamma$, the model parameters are updated as follows:

\begin{equation} \label{eqn:param_update}
\begin{split}
    &p = \begin{bmatrix} C_r & C_d & \beta_{elm,f}^T & \beta_{elm,r}^T \end{bmatrix}^T \\
    &v(t) = \gamma v(t-1) + (1-\gamma) \frac{\partial L}{\partial p} \\
    &p(t) = p(t-1) - \alpha v(t)
\end{split}
\end{equation}

We use a mini-batch with the size $K_{batch}$ for a robust update.


\subsubsection{Adaptive speed planning}
Along with updating the vehicle model, what also becomes crucial is to update the reference speeds along the race line. For example, if the friction coefficient reduces, the vehicle can only complete a turn successfully at lower cornering speeds. This requires reducing the reference speeds along the whole track accordingly. So, we generate the racing line speeds offline (c.f. Section \ref{subsec:racing_line}) and use the predicted value of the coefficient of friction for adaptive speed planning, i.e., assigning reference speed online according to time-varying friction. The friction coefficient dictates the maximum ratio of the force exerted to the normal force. Hence, we can obtain the predicted friction coefficient as the ratio of the maximum force exerted to the available normal force, which remains almost constant. As we predict the tire slip vs. force curve, we can obtain maximum force as the maximum magnitude of predicted values from ELM with slip angle inputs ranging from the observed slip angle values. Hence, the predicted friction coefficient is calculated as:


\begin{equation} \label{eqn:mu_pred}
    \begin{split}
        &\mu_{pred} = \frac{1}{m} \bigg(\max_{\alpha_f \in [\alpha_{f,min},\alpha_{f,max}]} | ELM_f(\alpha_f) | +\\ 
        &\ \ \ \ \ \ \ \ \ \ \max_{\alpha_r \in [-\alpha_{r,min},\alpha_{r,max}]} | ELM_r(\alpha_r)|\bigg) \\
    \end{split}
\end{equation} 
where $\alpha_{(f/r),min}$ and $\alpha_{(f/r),max}$ are the min and max values of the slip angles observed at any of the tires as of the current time step.





The optimization for control is formulated as follows:

\begin{equation} \label{eqn:opt}
    \begin{split}
        \min_{u_0,...u_{n-1}} &\sum_{k=1}^N \begin{Vmatrix} x_k - x_{ref,k}  \\ y_k-y_{ref,k} \end{Vmatrix}_Q + \sum_{k=1}^{N-1} \begin{Vmatrix} d_k - d_{k-1}  \\ \Delta \delta_k \end{Vmatrix}_R + \|\epsilon_k\|_S \\
        \text{s.t.  } &x_{k+1} = f_{corr} (x_k,u_k) \\
        \text{           } &x_0 = x(t) \\
        \text{           } &x_0 = x(t) \\
        \text{           } &A_k \begin{bmatrix}
            x_{k+1} \\
            y_{k+1}
        \end{bmatrix} \le b_k + \epsilon_k \\
        \text{           } &d_{min} \le d_k \le d_{max} \\
        \text{           } &\delta_{min} \le \delta_k \le \delta_{max} \\
        \text{           } &\Delta \delta_{min} \le \Delta \delta_k \le \Delta \delta_{max}
    \end{split}
\end{equation}
Here, the reference trajectory is obtained by sampling points on the target racing line at distances according to reference velocities at those points, similar to \cite{JainBayesRace2020}. The positive semi-definite matrices $Q$, $R$ and $S$ characterize the tracking, actuation and slack penalties, respectively. Slack variable $\epsilon_k$ is introduced to prevent infeasibilites. $f_{corr} = f_{e-kin} + e_{pred}$ is the corrected model, which outputs the next state. We set $f_{corr} = f_{e-kin}$ for $t \le N$ to train the models initially. $d_{min}, d_{max}, \delta_{min}, \delta_{max}$ are the actuation limits and $\Delta \delta_{min}, \Delta \delta_{max}$ are the steering rate limits. The final algorithm is described in Algorithm \ref{alg:algorithm}. We first run the e-kinematic model to train the dynamic model parameters for $K$ iterations, then we use the trained dynamic model to run the car while also updating the vehicle model with more collected data and adapt according to any changes in the model parameters. The reference position and speeds are obtained based on the current estimates of $\mu$, see Eq. \ref{eqn:mu_pred}. 

\begin{algorithm}[htbp]
    \SetKwInput{KwInput}{Input}                
    \SetKwInput{KwOutput}{Output}              
    \DontPrintSemicolon
    \LinesNumbered
    \SetKwFunction{FnSampleInitialState}{sample-initial-state}
    \SetKwFunction{FnQPController}{controller}
    \SetKwFunction{FnACC}{dynamics}
    \SetKwFunction{FnELM}{estimator}
    \SetKwFunction{FnELMu}{estimator.update}
    \SetKwFunction{FnLoss}{loss}
    \SetKwFunction{FnPlanner}{path-planner}
    \SetKwFunction{FnMu}{calc-mu}
    
    \KwInput{learning rate $\alpha$, momentum $\gamma$, Initial friction coefficient $\mu_{init}$, model parameters $m, l_f, l_r, I_z$, initialized estimators $ELM_{f}$, $ELM_{r}, C_r, C_d$, final time $t_f$, time step $dt$, offline training time threshold $t_{th}$.}
    \KwOutput{Optimized estimators and online control $k(x)$ at each time step}
    dataset $\xleftarrow{} \Phi$ \\
    \For{t = 0 \KwTo $t_f/dt$ }{
        $x_{t+1} \xleftarrow{} \FnACC(x_t,u_t)$ \tcp*{Get new state}
        $e_{(x_t,u_t)} \xleftarrow{} x_{t+1} - f_{e-kin}(x_t,u_t)$ 
        dataset.append($e_{(x_t,u_t)}$) \tcp{Collect data} 
        $L \xleftarrow{} \FnLoss(dataset[max(0,t-K_{batch}),t])$ \tcp*{Eqn \ref{eqn:loss}}
        $p \xleftarrow{} \FnELMu(L,\alpha,\gamma)$ \tcp*{Eqn \ref{eqn:param_update}}
        \If{$t > t_{th}$}{
            $f_{corr} \xleftarrow{} f_{e-kin} + e_{pred,p}$\;
            $\mu_{pred} \xleftarrow{} \FnMu(ELM_f,ELM_r)$ \tcp*{Eqn \ref{eqn:mu_pred}}
        }
        \Else{
            $f_{corr} \xleftarrow{} f_{e-kin}$ \;
            $\mu_{pred} \xleftarrow{} \mu_{init}$
        }
        $X_{ref} \xleftarrow{} \FnPlanner(x_{t+1},\mu_{pred})$
        $u_{t+1} \xleftarrow{} \FnQPController(X_{ref}, x_{t+1})$ \tcp*{Eqn \ref{eqn:opt}}
    }
    \caption{Online Learning based Adaptive Controller}
    \label{alg:algorithm}
\end{algorithm}

\section{RESULTS} \label{sec:results}

We perform experiments on the 1:43 numeric simulator and Carla high-fidelity vehicle simulator \cite{Dosovitskiy17}.

\subsection{Numeric simulator} \label{subsec:num_sim_results}
We use the 1:43 autonomous open-source race car simulation \cite{Liniger2015OptimizationbasedAR} to validate our algorithm. We first demonstrate our results on the ETHZ track \cite{Liniger2015OptimizationbasedAR}. For the numeric simulator experiments, we set $\alpha=0.002, \gamma=0.9, \mu_{start} = 1.0, K_{batch} = 300, t_{th}=6.2s, n_h=40, N=50, Q=\begin{bmatrix} 1 & 0 \\ 0 & 1 \end{bmatrix}, R = \begin{bmatrix} 0.005 & 0 \\ 0 & 1 \end{bmatrix}$. 

\subsubsection{Wear and tear of the tire}
\begin{figure}[htbp]
\begin{subfigure}{.24\textwidth}
    \centering
    \includegraphics[width=\textwidth]{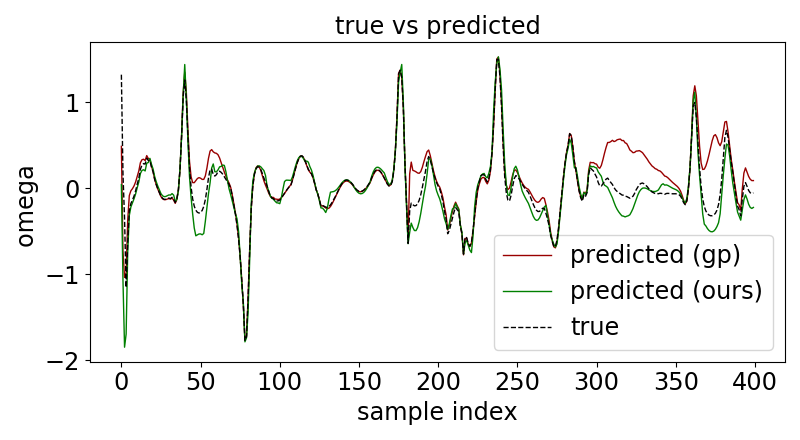}
    \caption{}
\end{subfigure}
\begin{subfigure}{.24\textwidth}
    \centering
    \includegraphics[width=\textwidth]{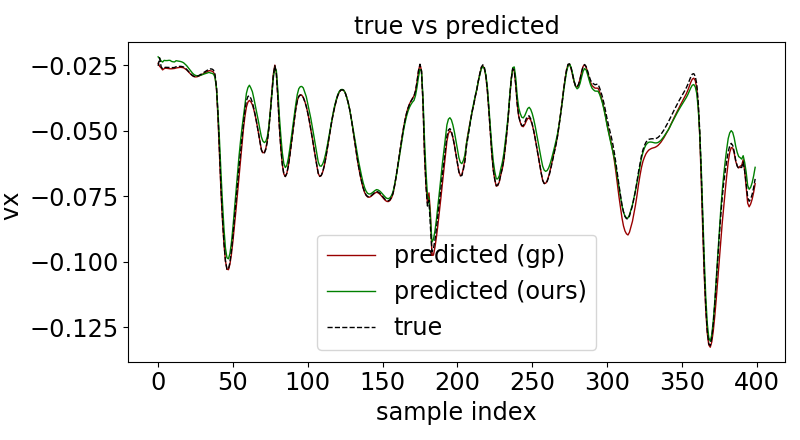}
    \caption{}
\end{subfigure}
\begin{subfigure}{.24\textwidth}
    \centering
    \includegraphics[width=\textwidth]{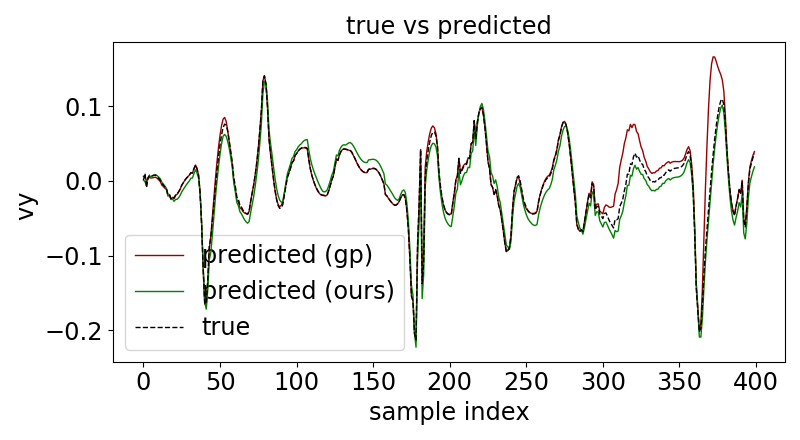}
    \caption{}
\end{subfigure}
\caption{Offline validation. Predictions of (a) $e_{v_x}$, (b) $e_{v_y}$ (c) $e_{\omega}$ using GP and Ours}
\label{fig:offline_val}
\end{figure}

We first show that our model is able to learn from offline data to draw a comparison with \cite{JainBayesRace2020}, which used GP to estimate the difference between dynamic and e-kinematic models. Our model is able to train quite well on the same collected data from the ETHZ track, as shown by the fitted errors in state predictions $v_x$, $v_y$ and $w$ in Fig. \ref{fig:offline_val}. Fig. \ref{fig:offline_val} also shows that our trained model gives quite comparable results to GP. Note that the computation time when the model is used for optimization with MPC is $0.04-0.07s$, while the GP requires computation times as long as $0.25-0.3s$ under the same condition. This clearly demonstrates ELM's improved computation efficiency over GP while yielding comparable accuracy. 

\begin{figure}[htbp]
\begin{subfigure}{.26\textwidth}
    \centering
    \includegraphics[width=\textwidth]{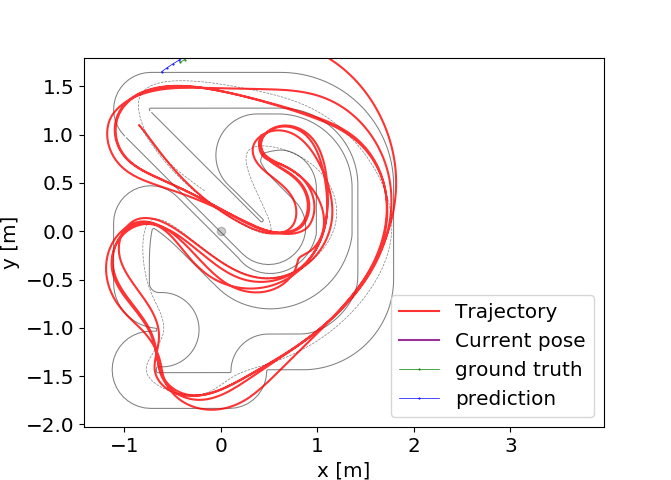}
    \caption{Trajectory}
\end{subfigure}
\begin{subfigure}{.22\textwidth}
    \centering
    \includegraphics[width=\textwidth]{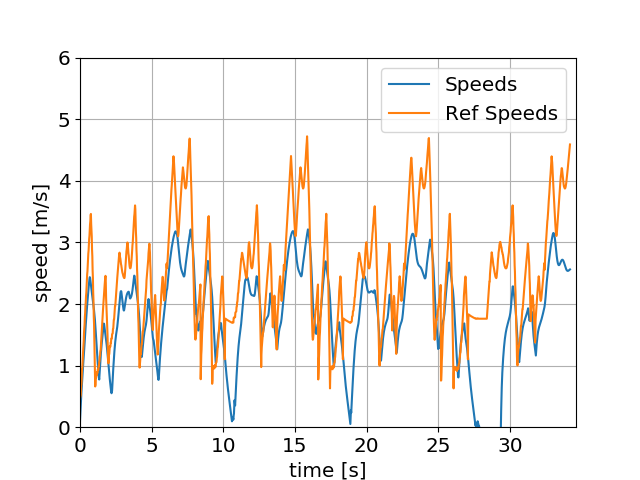}
    \caption{Speeds}
\end{subfigure}
\caption{Online run without model compensation}
\label{fig:without}
\end{figure}

\begin{figure}[htbp]
\begin{subfigure}{.26\textwidth}
    \centering
    \includegraphics[width=\textwidth]{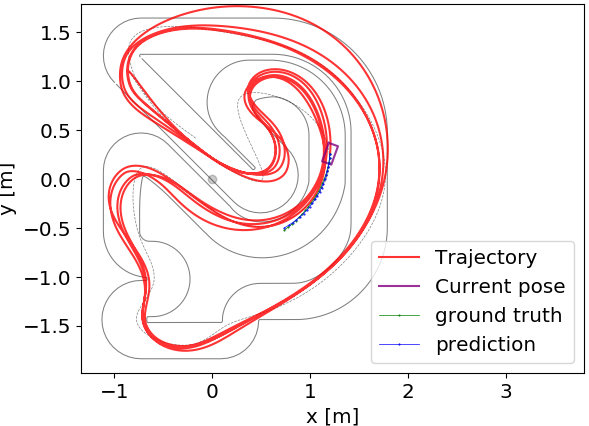}
    \caption{Trajectory}
\end{subfigure}
\begin{subfigure}{.22\textwidth}
    \centering
    \includegraphics[width=\textwidth]{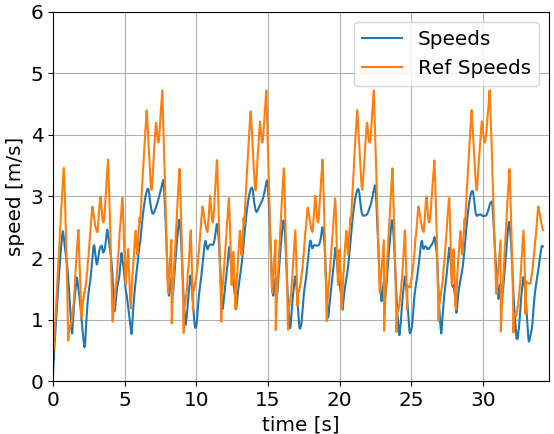}
    \caption{Speeds}
\end{subfigure}
\caption{Online run with model compensation, without adaptive speed planning}
\label{fig:with_const_speeds}
\end{figure}

\begin{figure}[htbp]
\begin{subfigure}{.26\textwidth}
    \centering
    \includegraphics[width=\textwidth]{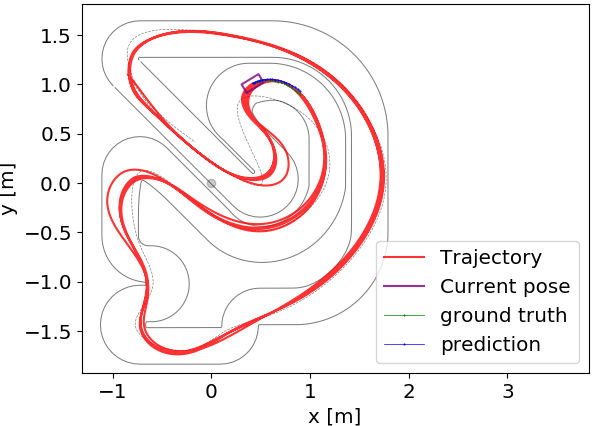}
    \caption{Trajectory}
\end{subfigure}
\begin{subfigure}{.22\textwidth}
    \centering
    \includegraphics[width=\textwidth]{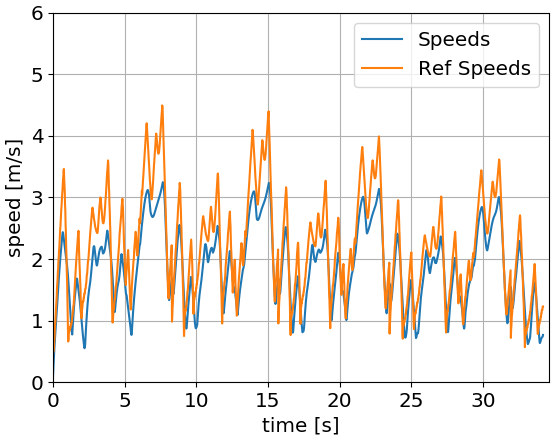}
    \caption{Speeds}
\end{subfigure}
\begin{subfigure}{.25\textwidth} 
    \centering
    \includegraphics[width=\textwidth]{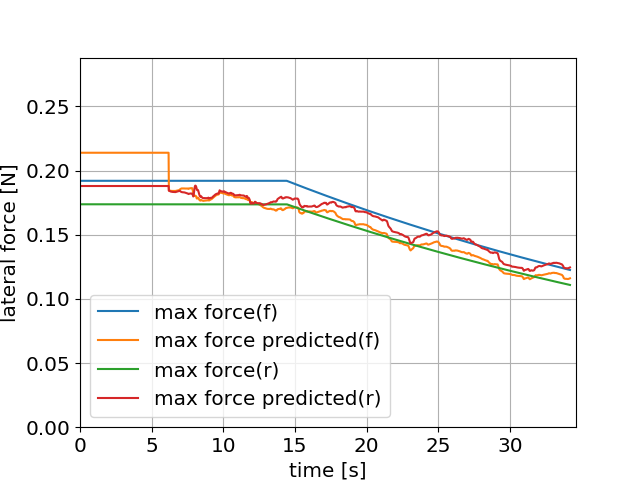}
    \caption{$\mu$ predictions}
    \label{subfig:mu_changes}
\end{subfigure} 
\caption{Online run with model compensation \& adaptive speed planning}
\label{fig:with_var_speeds}
\end{figure}

\begin{figure}[htbp]
\begin{subfigure}{.24\textwidth}
    \centering
    \includegraphics[width=\textwidth]{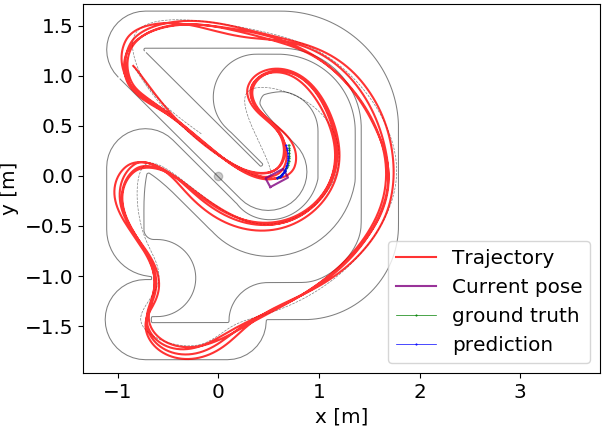}
    \caption{}
    \label{subfig:with_gp}
\end{subfigure}
\begin{subfigure}{.24\textwidth}
    \centering
    \includegraphics[width=\textwidth]{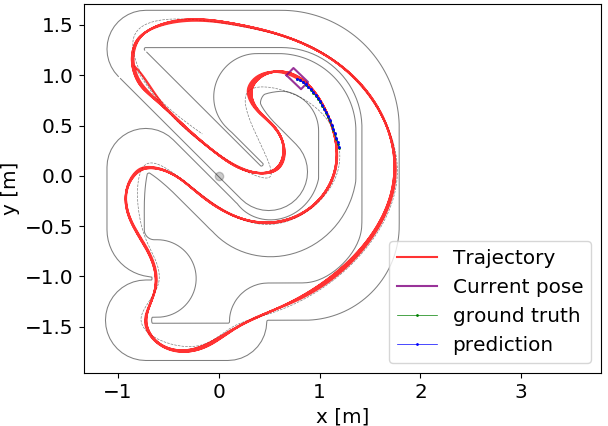}
    \caption{}
    \label{subfig:with_oracle}
\end{subfigure}
\caption{Trajectories for Online run with model compensation (a) Using GP (b) Oracle }
\label{fig:with_gp_oracle}
\end{figure}

We also compare path-following respectively without and with the use of adaptive model compensation in Fig. \ref{fig:without} and Fig. \ref{fig:with_var_speeds}. We modify the actual $\mu$, i.e., change parameters $D_f$ and $D_r$, as shown in Fig. \ref{subfig:mu_changes}, and the predicted values as well. As can be seen, using our approach with adaptive model compensation clearly yields better performance. Using adaptive speed planning also allows the vehicle to stay within lane boundaries. If speed references are not changed, as can be observed in Fig. \ref{fig:with_const_speeds}, the vehicle is still able to complete its maneuvers, but leaves the track several times. This is because even with a good updated model, unadjusted reference speeds push the vehicle beyond its limits at turns. Not updating the vehicle model leads to the vehicle completely failing and losing control at a region on the $3^{rd}$ lap where it had to completely stop to gain control. Finally, we also compare against an oracle (Fig. \ref{subfig:with_oracle}), where we use the exact Pacejka friction parameters and make the run with the same parameters. As can be observed, the trajectory followed with our method (Fig. \ref{fig:with_var_speeds}.a) and the lap times are almost the same as those of the oracle (Fig. \ref{subfig:with_oracle}).

Moreover, we also compare against the use of adaptive Gaussian processes, where we adjust the speeds according to the ground truth value of $\mu$ (see Fig. \ref{subfig:mu_changes}), and this yields slightly worse (very close) results than ours but with much higher computation times. Also, as can be observed from Fig. \ref{fig:with_var_speeds} and Fig. \ref{subfig:with_gp}, the trajectory followed when using GP goes further away from the reference with decreasing $\mu$ while our method, the trajectory remains nearly the same similar to the oracle (see Fig. \ref{subfig:with_oracle}). It should also be noted that if using GP requires a separate $\mu$ prediction process to replace and obtain an estimate close to the ground truth value, while in our method we inherently obtain the predicted value of $\mu$ according to Equation \ref{eqn:mu_pred}. 

We compare some of the run statistics in Table \ref{tab:num_results}. As can be seen: 1) our lap time is slightly worse than the oracle but the safety (i.e., the violation time) has been significantly improved; 2) the mean deviation from the racing line is least when our method is used with adjusted speeds (close to the oracle); 3) our approach outperforms the state-of-the-art GP-based controller \cite{JainBayesRace2020} (when used in online setting) in terms of every criterion as well as the computation efficiency. The computation times for our method are within $0.05s\pm0.02s$ and for GP is $0.3s\pm0.05s$ when used online on our machine (AMD Ryzen 7 5000 series CPU, 16 GB RAM).

\begin{table}[]
\setlength\tabcolsep{2pt}
\begin{tabular}{|l|l|l|l|l|l|l|}
\hline
Method                                                                          & \begin{tabular}[c]{@{}l@{}}Lap 0\\ time\end{tabular} & \begin{tabular}[c]{@{}l@{}}Lap 1\\ time\end{tabular} & \begin{tabular}[c]{@{}l@{}}Lap 2\\ time\end{tabular} & \begin{tabular}[c]{@{}l@{}}Lap 3\\ time\end{tabular} & \begin{tabular}[c]{@{}l@{}}Mean dev.\\ from rac-\\ ing line\end{tabular} & \begin{tabular}[c]{@{}l@{}}Violation\\ time\end{tabular} \\ \hline
\begin{tabular}[c]{@{}l@{}}Without\\ adaptation\end{tabular}                    & 8.44s                                                & 8.34s                                                & 8.72s                                                & -                                                    & 0.141m                                                                   & 4.32s                                                    \\ \hline
\begin{tabular}[c]{@{}l@{}}With\\ adaptation\\ (Constant\\ speeds)\end{tabular} & 8.26s                                                & 7.26s                                                & 7.56s                                                & 8.14s                                                & 0.1104m                                                                  & 2.84s                                                    \\ \hline
\begin{tabular}[c]{@{}l@{}}With\\ adaptation\\ (Var speeds, \textbf{ours})\end{tabular}        & 8.28s                                                & 7.42s                                                & 7.76s                                                & 8.46s                                                & 0.0832m                                                                  & 0.46s                                                    \\ \hline
GP                                                                              & 8.28s                                                & 7.44s                                                & 7.84s                                                & 8.56s                                                & 0.0989m                                                                  & 0.62s                                                    \\ \hline
Oracle                                                                          & \textit{7.92s}                                       & \textit{7.40s}                                       & \textit{7.72s}                                       & \textit{8.40s}                                       & \textit{0.0511m}                                                         & \textit{0s}                                              \\ \hline
\end{tabular}
\caption{Comparison in statistics for numeric simulator. Violation time is the total time for which the vehicle was outside the track limits}
\label{tab:num_results}
\end{table}

\subsubsection{Sudden friction change}
\begin{figure}[htbp]
\begin{subfigure}{.26\textwidth}
    \centering
    \includegraphics[width=\textwidth]{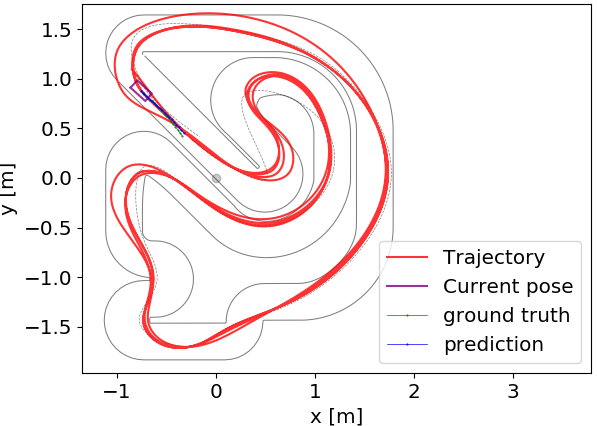}
    \caption{Trajectory}
\end{subfigure}
\begin{subfigure}{.22\textwidth}
    \centering
    \includegraphics[width=\textwidth]{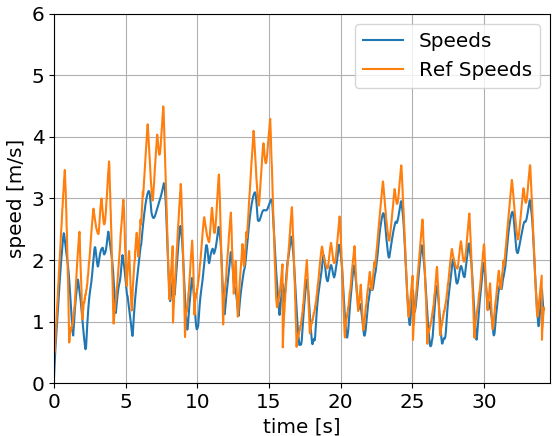}
    \caption{Speeds}
\end{subfigure}
\begin{subfigure}{.3\textwidth} 
    \centering
    \includegraphics[width=\textwidth]{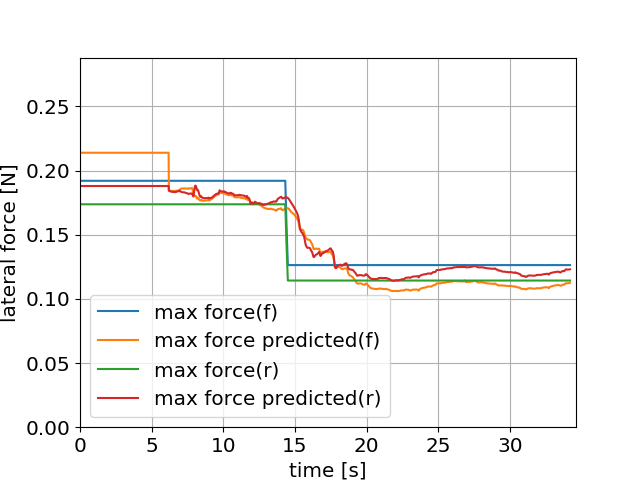}
    \caption{$\mu$ predictions}
    \label{subfig:mu_changes_sudden}
\end{subfigure} 
\caption{Online run with model compensation \& adaptive speed planning (sudden friction change experiment)}
\label{fig:with_var_speeds_sudden}
\end{figure}

\begin{figure}[htbp]
\begin{subfigure}{.26\textwidth}
    \centering
    \includegraphics[width=\textwidth]{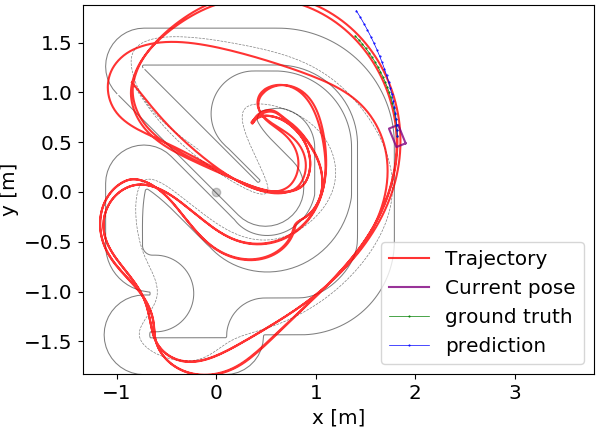}
    \caption{Trajectory}
\end{subfigure}
\begin{subfigure}{.22\textwidth}
    \centering
    \includegraphics[width=\textwidth]{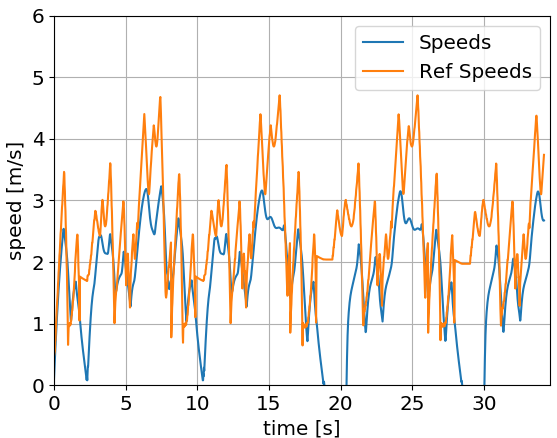}
    \caption{Speeds}
\end{subfigure}
\caption{Online run without model compensation (sudden friction change experiment)}
\label{fig:without_sudden}
\end{figure}

Now, we test our approach against a situation when the friction coefficient suddenly changes. This is possible, for example, when a sudden heavy rain changes the road's friction coefficient. For this case, we demonstrate the path followed by our approach (See Fig. \ref{fig:with_var_speeds_sudden}) and compare it to when the model compensation is not considered (See Fig. \ref{fig:without_sudden}). As observed, with model compensation, the vehicle is able to estimate the change in friction coefficient quite well and abruptly. It does cross the outer boundary at the beginning but is able to adjust the speeds to stay safe within the lane limits later, whereas if compensation is not considered, the vehicle completely loses control at several points after $\mu$ changes.

\subsection{Carla}
\begin{figure}[!htbp]
\begin{subfigure}{.24\textwidth}
    \centering
    \includegraphics[width=\textwidth]{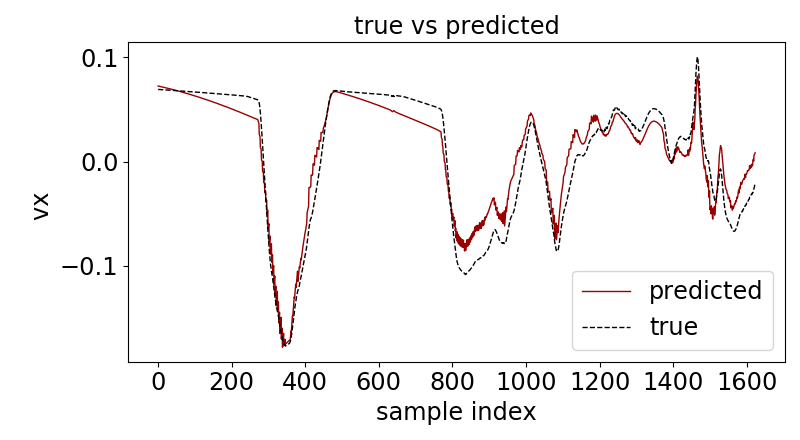}
    \caption{}
\end{subfigure}
\begin{subfigure}{.24\textwidth}
    \centering
    \includegraphics[width=\textwidth]{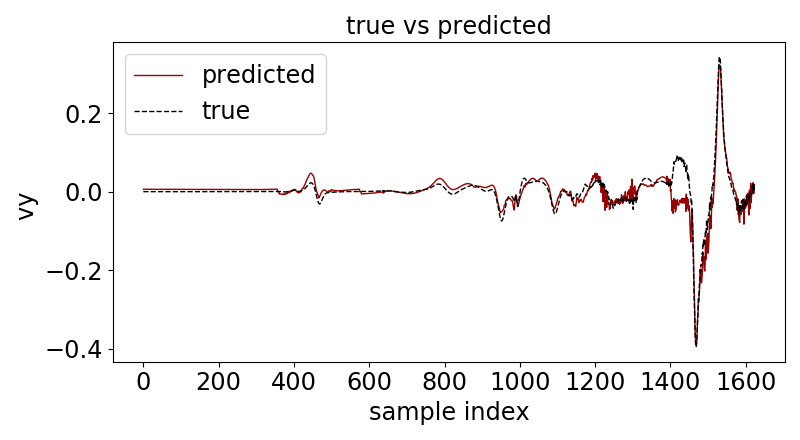}
    \caption{}
\end{subfigure}
\begin{subfigure}{.24\textwidth}
    \centering
    \includegraphics[width=\textwidth]{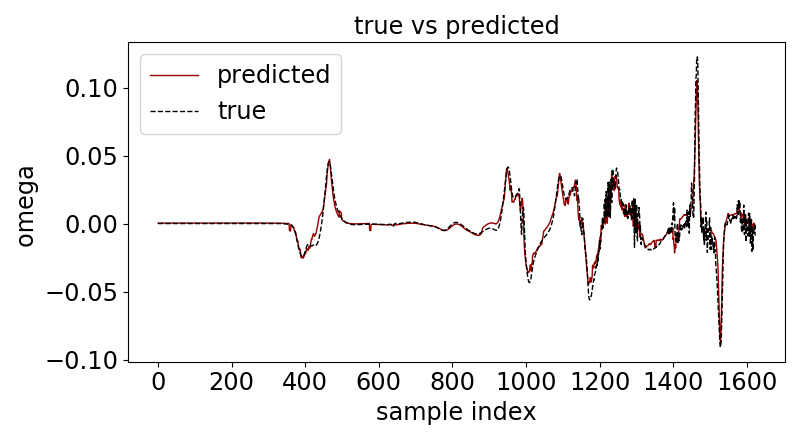}
    \caption{}
\end{subfigure}
\begin{subfigure}{.22\textwidth}
    \centering
    \includegraphics[width=\textwidth]{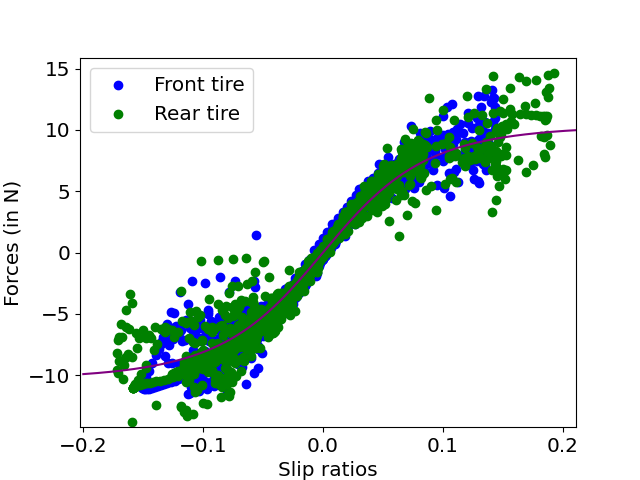}
    \caption{}
\end{subfigure}
\caption{Offline validation (Carla). Predictions of (a) $e_{v_x}$, (b) $e_{v_y}$ (c) $e_{\omega}$ and (d) observed tire force predictions vs. predicted values}
\label{fig:offline_val_carla}
\end{figure}

\begin{figure}[!htbp]
\begin{subfigure}{.4\textwidth}
    \centering
    \includegraphics[width=\textwidth]{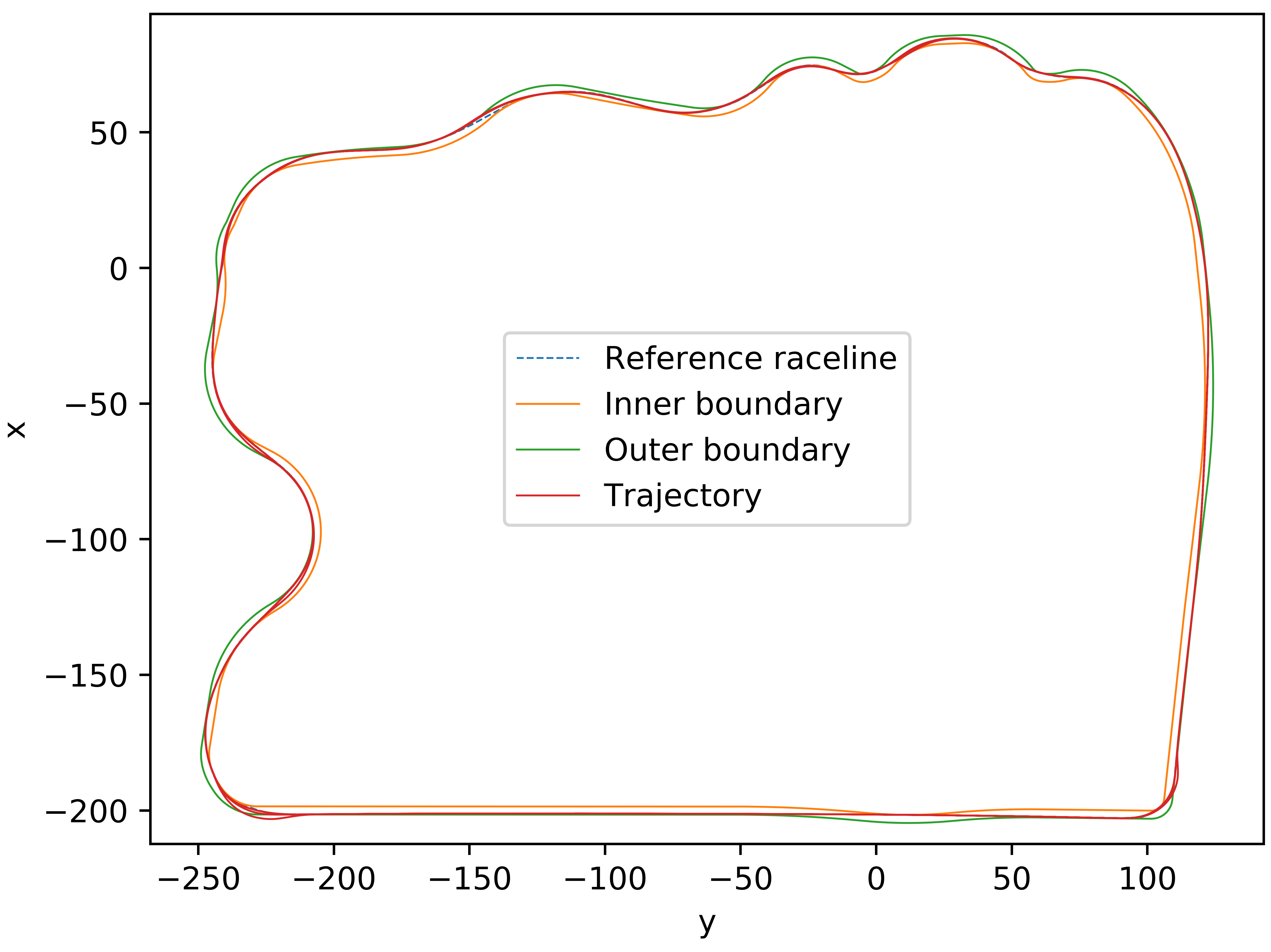}
    \caption{Trajectory}
\end{subfigure}
\begin{subfigure}{.24\textwidth}
    \centering
    \includegraphics[width=\textwidth]{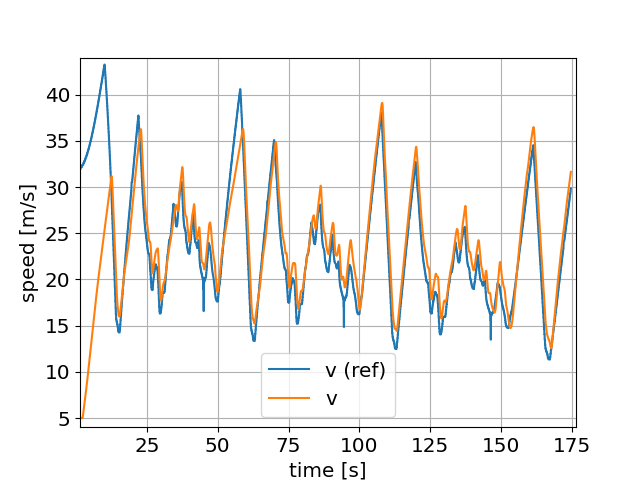}
    \caption{Speeds}
\end{subfigure}
\begin{subfigure}{.24\textwidth} \label{subfig:mu_changes_carla}
    \centering
    \includegraphics[width=\textwidth]{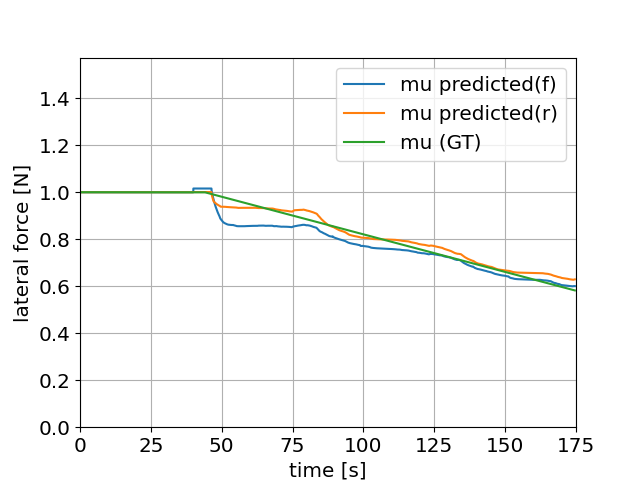}
    \caption{$\mu$ predictions}
\end{subfigure} 
\caption{Online run with model compensation in Carla}
\label{fig:with_carla}
\end{figure}

\begin{figure}[!htbp]
\begin{subfigure}{.4\textwidth}
    \centering
    \includegraphics[width=\textwidth]{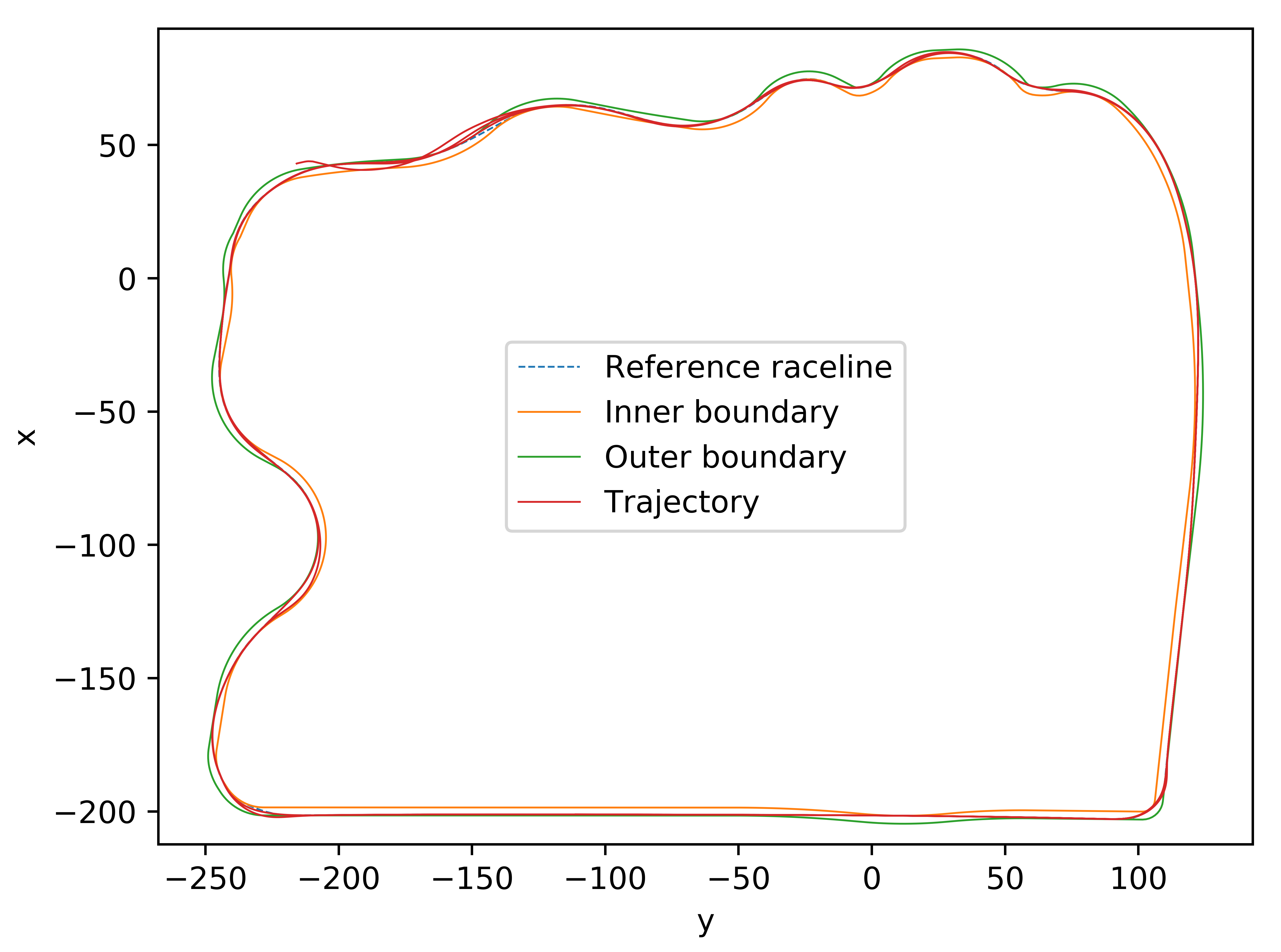}
    \caption{Trajectory}
\end{subfigure}
\begin{subfigure}{.24\textwidth}
    \centering
    \includegraphics[width=\textwidth]{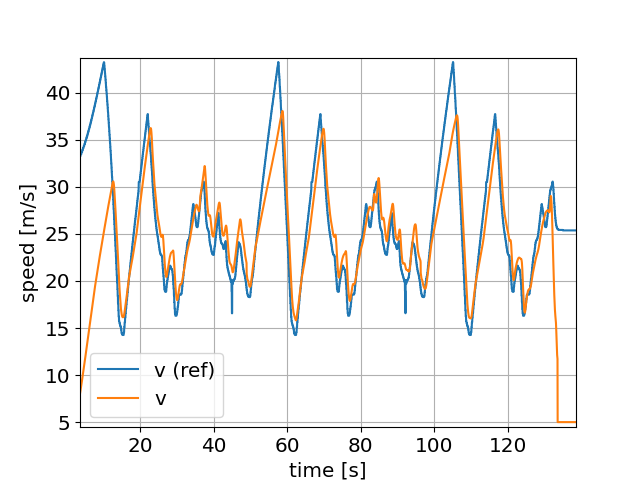}
    \caption{Speeds}
\end{subfigure}
\caption{Online run without model compensation in Carla}
\label{fig:without_carla}
\end{figure}

Finally, we perform experiments in Carla \cite{Dosovitskiy17} with time-varying tire friction parameters to test our online adaptation algorithm. We first obtain the racing line on the Carla track, as depicted in Fig. \ref{fig:racing_line_carla_}. For offline validation, we check the lateral and longitudinal forces from the trained model on one lap of data and compare them to the lateral and longitudinal force values obtained from the simulator. As shown in Fig. \ref{fig:offline_val_carla}, although these values are noisy, our trained model archives high fitting accuracy. For the online run, we set $\alpha=0.002, \gamma=0.9, \mu_{start} = 1.0, K_{batch} = 2000, t_{th}=44s, n_h=40, N=50, Q=\begin{bmatrix} 0.1 & 0 \\ 0 & 0.1 \end{bmatrix}, R = \begin{bmatrix} 0.005 & 0 \\ 0 & 1 \end{bmatrix}$. The online run with friction decay is shown in Fig. \ref{fig:with_carla}. Both tires have the same tire model. For comparison, we show the results with and without adaptive model compensation (See Fig. \ref{fig:with_carla} and Fig. \ref{fig:without_carla}). As observed, without adaptive model compensation, the vehicle moves out of bounds after about $60s$ once the friction coefficient starts reducing and collides with an obstacle.


\vspace{-5mm}



\section{CONCLUSION and FUTURE WORK} \label{sec:conclusion}

We present a learning-based control algorithm that both significantly reduces the effort required for system identification and allows online adaptation. Starting with a simple geometric model we collect training data which are used by our algorithm to iteratively correct the mismatch between the geometric and the actual vehicle model, and also adapt to any changes. This allows racing algorithms to devise an aggressive strategy without worrying about the vehicle model parameters and wear and tear of the tires. We demonstrate our performance in the presence of simulated wear and tear in both the 1:43 scale autonomous racing platform \cite{Liniger2015OptimizationbasedAR} and the Carla Simulator \cite{Dosovitskiy17}. In the future it would be interesting to explore using uncertainty in the trained models, which can be used in the robust controller formulation. It would also be interesting to explore how we can guide the training effectively to push the vehicle beyond it's known limits to explore it's limits and improve the model's confidence towards it. This can be done using a similar approach to adaptive $\epsilon-$greedy Q-learning \cite{DOSSANTOSMIGNON20171146}, where $\epsilon$ can be changed according to the vehicle's current safety index.    

\bibliographystyle{IEEEtran}
\bibliography{./IEEEfull,refs}

\end{document}